\documentclass[letterpaper, 10 pt, conference]{ieeeconf}  

\usepackage{times}
\let\labelindent\relax
\usepackage{enumitem}
\usepackage{graphicx}
\usepackage{multicol}
\usepackage{bm}
\usepackage{algorithm}
\usepackage{algorithmic}
\usepackage{siunitx}
\usepackage{balance}

\usepackage{bbding}
\usepackage{pifont}
\usepackage{wasysym}
\usepackage{amssymb}
\usepackage{color,soul}
\usepackage{url}
\usepackage{caption}

\usepackage{tabularx}
   \newcolumntype{C}{>{\centering\arraybackslash}X}
   \newcolumntype{L}{>{\raggedright\arraybackslash}X}
   \newcolumntype{R}{>{\raggedleft\arraybackslash}X}

\newcommand{\colornewparts}[0]{\color{black}}

\IEEEoverridecommandlockouts                              
\makeatletter
\let\NAT@parse\undefined
\makeatother
\usepackage[numbers]{natbib}

\title{\LARGE \bf
Probabilistic Visual Navigation with Bidirectional Image Prediction
}

\author{Noriaki Hirose$^{1}$, Shun Taguchi$^{1}$, Fei Xia$^{2}$, Roberto Mart\'{i}n-Mart\'{i}n$^{2}$,\\ Kosuke Tahara$^{1}$, Masanori Ishigaki$^{1}$, and Silvio Savarese$^{2}$
\thanks{$^{1}$N. Hirose, {\it et al.} are with TOYOTA Central R$\&$D Labs., INC, Japan
        {\tt\small hirose@mosk.tytlabs.co.jp}}%
\thanks{$^{2}$F. Xia, {\it et al.} are with the Department of Computer Science, Stanford University,
        CA 94305, USA
        {\tt\small feixia@stanford.edu}}%
}

\begin{document}

\maketitle
\thispagestyle{empty}
\pagestyle{empty}

\begin{abstract}
Humans can robustly follow a visual trajectory defined by a sequence of images (i.e. a video) regardless of substantial changes in the environment or the presence of obstacles. We aim at endowing similar visual navigation capabilities to mobile robots solely equipped with a RGB fisheye camera. We propose a novel probabilistic visual navigation system that learns to follow a sequence of images with bidirectional visual predictions 
conditioned on possible navigation velocities. 
By predicting bidirectionally (from start towards goal and vice versa) our method extends its predictive horizon enabling the robot to go around unseen large obstacles that are not visible in the video trajectory. Learning how to react to obstacles and potential risks in the visual field is achieved by imitating human teleoperators. Since the human teleoperation commands are diverse, we propose a probabilistic representation of trajectories that we can sample to find the safest path. 
We evaluate our navigation system quantitatively and qualitatively in multiple simulated and real environments and compare to state-of-the-art baselines.
Our approach outperforms the most recent visual navigation methods with a large margin with regard to goal arrival rate, subgoal coverage rate, and success weighted by path length (SPL). Our method also generalizes to new robot embodiments never used during training.
\end{abstract}
\section{Introduction}
%


Imagine you are tasked with reaching a location described by an image in a novel environment. This task can range from difficult to impossible, depending on the size and aspect of the environment. Now imagine you are given the same task but also a sequence of images depicting what you would see if you would navigate correctly to the goal location. This is a task that most humans are able to perform based only on visual information. In this work we propose a mechanism to enable robots to reach a destination by following images. 

Previous approaches have studied the problem of robot navigation following a sequence of images~\cite{chaumette2006visual, chaumette2007visual, NIPS2018_7357}. However, most of these methods suffer if the environment changes from when the sequence of images was acquired, or if they have to deviate from the path to avoid a collision. Compared to robots, most humans can follow a visual trajectory~(VT) even if the environment changed significantly due to changes in illumination, objects that moved or people or other dynamic agents that are placed differently in the scene~\cite{wang2002human}. They can also robustly deviate from the path to avoid obstacles. To do that, cognitive scientists argue that humans perform mental operations on predictive visual models~\cite{chersi2013mental, arnold2016mental}. Taking inspiration from these insights we developed a navigation algorithm that generates virtual future images based on possible navigation strategies, and matches these images to the current view so as to find the best path to move safely. Concretely, our method will search for optimal navigation velocities to progress along a sequence of images towards a goal, even when there are significant changes in the environment, while keeping the robot safe from colliding with obstacles and other risks in the environment.

\begin{figure}[t]
  \begin{center}
  \hspace*{-5mm}
      \includegraphics[width=0.85\hsize]{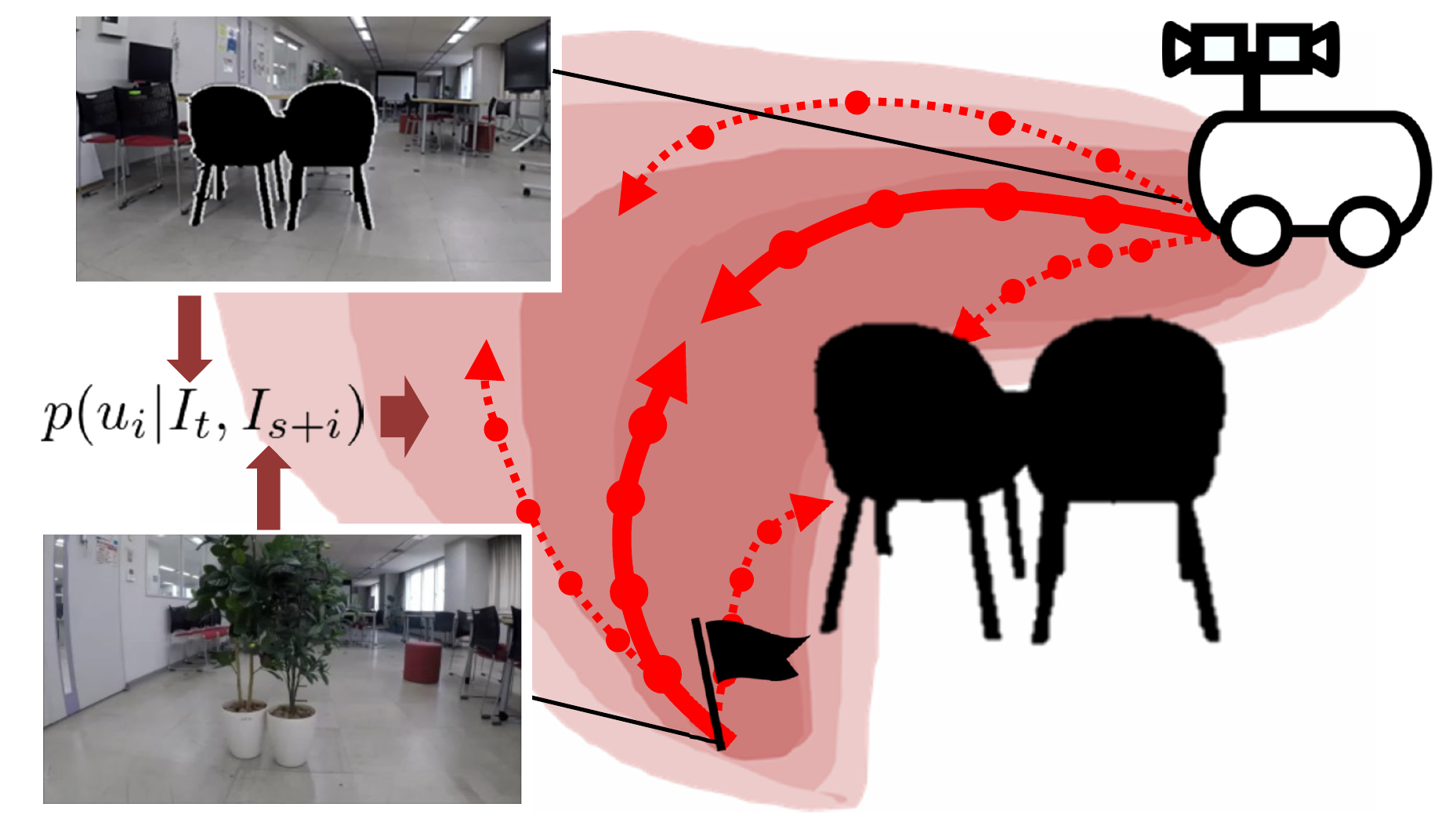}
  \end{center}
      \vspace*{-3mm}
	\caption{\small Our method allows the robot to avoid unseen large obstacles and navigate towards a goal given by an image using bidirectional image prediction from both A) current image $I_t$ and B) the goal image $I_{s+i}$. A probabilistic representation of the distribution of human teleoperation demonstrations ($p(u_i|I_t, I_{s+i})$, red curves) is used to sample multiple collision avoidance trajectories to select safe and reachable one.}
  \label{f:pull_fig}
  \vspace*{-7mm}
\end{figure}

Previous approaches have proposed comparable visual predictive models to find optimal commands for navigation and manipulation~\cite{pmlr-v78-drews17a, drews2019vision, hirose2019dvmpc, finn2017deep, kase2019learning}. 
These approaches have shown that collision avoidance for a large obstacle unseen in the VT is generally difficult because the obstacle occludes the view.
It may be difficult to recognize the direction to move to reach the next image in the sequence.
In addition, longer visual prediction is needed to deviate largely from the VT to avoid large obstacles with spacial margin.
However, due to the increasing uncertainty in future states introduced by (stochastic) forward models, the predicted images often grow blurry with each step~\cite{ebert2018visual} hindering the search of an appropriate robot command. 

We propose to take inspiration from classical planning~\cite{lavalle2006planning}, and perform \textit{bidirectional search}, not in Cartesian but in image space. The main idea of bidirectional search is to perform the search from both the initial location and the goal locations until a connection is found~\cite{lavalle2001randomized} as shown in Fig.~\ref{f:pull_fig}. Our bidirectional image prediction is inspired by this method, and aims to improve search efficiency while alleviating the deterioration in the images due to long-horizon predictions.

In addition, to learn to avoid obstacles and come back to the VT, we propose to imitate human demonstrations. We asked human teleoperators to guide a mobile robot while we collected images. This allowed us to learn the correlation between obstacles in the image and navigation strategies and distill them into a model-predictive architecture. Since the navigation strategies from humans present large variability (e.g., between aggressive and conservative teleoperators, or when multiple ways to avoid an obstacle are possible) we encode the possible trajectories with a probabilistic representation. This representation can be used to sample multiple navigation strategies towards a visual goal and select the best one, based on a visual-based traversability method~\cite{hirose2018gonet}.

The main contributions are summarized as follows:
\begin{itemize}[leftmargin=*]
\item We propose a novel bidirectional image prediction method to {\colornewparts predict images given possible virtual velocities} from both current and desired visual states enabling {\colornewparts avoidance of large obstacles occludes the view.}
\item We developed a probabilistic representation of possible future trajectories learned from human teleoperators.
\item {\colornewparts We combined these components into a visual model predictive architecture that allows the robot to navigate towards a goal, following a sequence of images robustly.} 
\end{itemize}

We evaluate our method and compare it with various baselines in  photo-realistic simulated environments and the real world.
In total, we performed 13000 trials for about 572 hours in the simulator and 80 trials in the real environment.
We test the robot's behavior during navigation in challenging environments containing various static obstacles, some that were not present when the VT was generated, as well as dynamic obstacles.
Moreover, to validate the capability of our method to transfer between different robot embodiments, we tested on a mobile robot that was never used to collect training data.
Even in these conditions, our method performs well as the robot can successfully reach the goal.



\section{Related Works}
In the domain of visual navigation, the methods can roughly be categorized into model-based visual navigation and learning based visual navigation, and we summarize works from both categories as below. 

\textbf{Model-Based Visual Navigation:}
Researchers have proposed solutions to visual navigation based on visual servoing and visual SLAM. Visual servoing~\cite{hutchinson1996tutorial,chaumette2006visual} controls an agent to minimize the difference between current state and a goal state.
As the difference is defined in image space, the performance suffers when the environment changes or large obstacles  occlude a large parts of the environment. Differently, our method copes with large obstacles using bidirectional predictions. 

Navigation methods based on visual SLAM~\cite{mur2017orb,kim2013perception,karlsson2005vslam} simultaneously perform robot localization and map building using camera images, and compute actions to achieve the goal. The success of visual SLAM-based methods relies on acquiring an accurate metric model and their performance decay due to failures in the mapping. Our method does not rely on any metric map and only requires an input trajectory defined by a sequence of images.

\textbf{Learning-Based Visual Navigation:} Recently, multiple learning-based approaches have been proposed for visual navigation. Topological visual navigation~\cite{savinov2018semiparametric,chen2019behavioral,meng2019scaling,chaplot2020neural} involves localization and planning from a topological representation of the environment that represents the connectivity between regions. Latest advances of reinforcement learning~\cite{Wijmans2020DD-PPO:,mishkin2019benchmarking,zhu2017target,kahn2018self} and imitation learning~\cite{codevilla2018end,pokle2019deep,kase2019learning} have also pushed the state of the art in visual navigation. We will focus the discussion on the two families of learning-based methods that are the closest to ours, visual predictive control methods and probabilistic control methods. 

%
%
Visual MPC~(Model Predictive Control) is used for manipulation and navigation. \citet{finn2017deep} proposed a visual MPC approach using an image-based predictive neural network model. \citet{pathak2018zero} proposed visual navigation with one-step image prediction from raw camera images. \citet{hirose2019dvmpc} proposed to train control policies for navigation by minimizing the visual MPC objective while penalizing untraversable areas. In these works, the predictive horizon is relatively short and thus, they cannot deviate far from the original path to avoid obstacles. Compared to these methods, our method is also computationally lighter and can be used for online navigation.




In the navigation domain, \citet{drews2019vision,pmlr-v78-drews17a} proposed a navigation system with image-based localization using a map and a control method based on non-visual MPC with sampling-based optimization for a small race car. \citet{amini2019variational} trained a probabilistic model of driver behavior using the Gaussian mixture model and simultaneous localization via a trained control policy. 

Similar to ours, \citet{NIPS2018_7357} proposed using a probabilistic approach for a robot to navigate along consecutive subgoal images between the start and final positions. However, their method shows oscillatory behaviors because their control commands are discrete (turn left, turn right, and go straight), while our method uses continuous control and generates smoother navigation trajectory. \citet{bansal2019-lb-wayptnav} also suggested using the visual MPC approach for navigation and collision avoidance with spline trajectories. Unlike our method, they can not take arbitrary collision avoidance motions and require the ideal goal position in Cartesian space instead of the goal view, which requires some metric mapping.




Compared with the most similar method, \citet{hirose2019dvmpc}, our work applies a probabilistic approach to understand non-deterministic teleoperator's behavior as the distribution of trajectory instead of single deterministic one.
In addition, we propose bidirectional prediction to recognize the direction to move to the next subgoal even if the large obstacles occlude the view of the next subgoal image. The bidirectional approach can also double the predictive horizon to allow large deviation from the VT to avoid the collision. For the navigation system we also incorporate visual localization to improve the navigation performance. 
As a result, our method can follow the VT to arrive at the goal even if large deviation from the VT is enforced by collision avoidance.
%
\begin{figure}[t]
  \begin{center}
  \vspace*{1mm}
      \includegraphics[width=0.98\hsize]{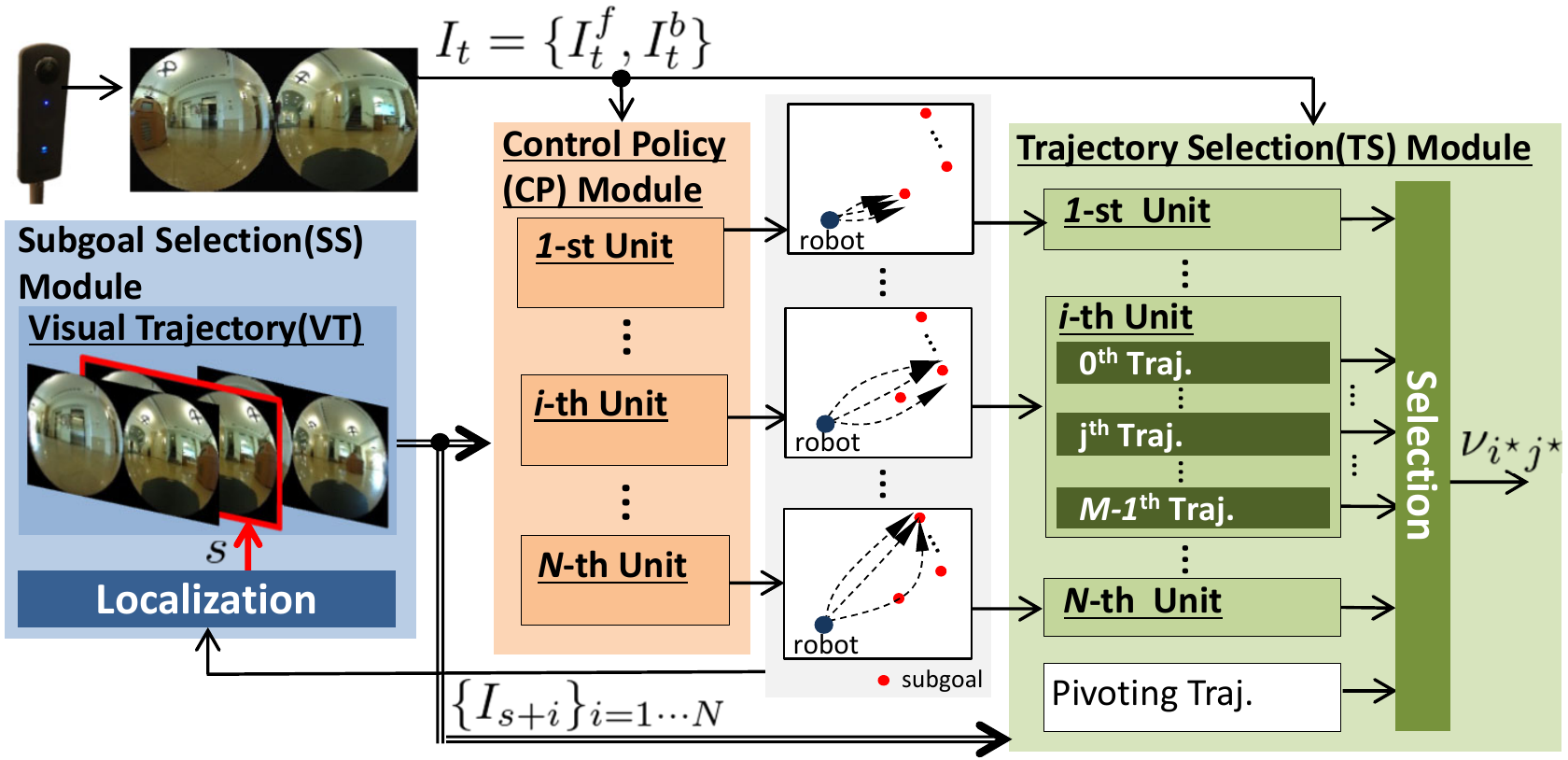}
  \end{center}
      \vspace*{-3mm}
	\caption{\small {\bf Proposed System:} Our method contains three main modules: a subgoal selection(SS) module (blue), a control policy(CP) module (orange), and a trajectory selection(TS) module (green). SS module determines the next subgoals to reach in the trajectory $\{ I_{s+i}\}_{i=1 \cdots N}$ and feeds them into CP and TS modules. Each unit in CP module generates $M$ sequences of velocities that would move the robot towards a subgoal location. TS module chooses the optimal sequence of velocities by scoring the reachability and traversability for the subgoals. If none of the generated trajectories is safe enough, the TS module outputs a pivoting trajectory~(90$^\circ$ rotation) that helps the robot to relocalize and continue moving.}
  \label{f:block_diagram}
  \vspace*{-7mm}
\end{figure}
%
\section{Method}
\label{s_method}

In this section we present a system to navigate using images so as to follow a pre-recorded VT. A VT is defined by a sequence of images (\textit{subgoal images}, $I_{sg}$) ending at a \textit{final goal image}, $I_g$. The location where a \textit{subgoal image} was acquired is called a \textit{subgoal location}. 
While our method is independent of the type of camera images in the implementation (Sec.~\ref{s_training} and Sec.~\ref{s_exp}) we will use images captured from a 360$^\circ$ RGB fisheye camera by concatenating front and back side fisheye images in the channel direction. And we use trajectories collected by teleoperating a robot.
Fig.~\ref{f:block_diagram} depicts an overview of our proposed probabilistic control system with bidirectional prediction. Our system is composed of three main modules, a subgoal selection~(SS), a control policy~(CP), and a trajectory selection~(TS) module, that we explain in the following.

The SS module (Fig~\ref{f:block_diagram}, blue, Sec.~\ref{ss_vlm}) stores the VT and infers the subgoal image, $I_s$, that was acquired from the subgoal location closest to the current robot location (indicated in red in Fig.~\ref{f:block_diagram}).
The $N$ consecutive images after $I_s$ in the VT, $\{ I_{s+i}\}_{i=1 \cdots N}$, are passed to both the CP and TS modules. Using these $N$ consecutive subgoal images, our method is able to transition to later subgoals even when the closest subgoals are occluded by large obstacles not present in the environment when the VT was acquired. This allows our method to deviate largely from the VT, executing maneuvers that will reach a further step towards the goal.

To process the $N$ consecutive subgoal images, we propose a CP module composed by $N$ units with shared weights (Fig~\ref{f:block_diagram}, orange, Sec.~\ref{ss_cpm}). 
All units use as input the current sensor image, $I_t$, and a different subgoal image $I_{s+i}$. 
The output of the unit is $M$ command trajectories $\nu_{i,j}$, with $j\in \{0\ldots M-1\}$.
A command trajectory $\nu_{i,j}$ is a sequence of $2L$ consecutive velocity commands, $\{{u}_{i, j, k} \}_{k=0 \cdots 2L-1}$, that executed at a constant time interval $\Delta_t$ would move the robot from its current location towards the subgoal location of $I_{s+i}$.
The $M$ trajectories correspond to samples in the  distribution of possible trajectories towards the subgoal, $p(\nu|I_t, I_{s+i})$, learned from a dataset of human teleoperation maneuvers.

Finally, the TS module (Fig.~\ref{f:block_diagram}, green, Sec.~\ref{ss_sm}) chooses the best command trajectory $\nu_{i^\star,j^\star}$ to execute, i.e., the best subgoal image to reach and the best sample trajectory to move towards it. The selection is based on the reachability of the subgoal and the traversability (safety) of the path. If none of the generated trajectories is safe enough, the TS module outputs a pivoting trajectory (90$^\circ$ rotation) that helps the robot to relocalize and continue the navigation. 
The trajectory is passed to a controller that executes the first velocity command, and the inference process starts again with a newly acquired sensor image $I_{t+\Delta_t}$. In the following, we will describe these three modules in detail.

\subsection{Control Policy~(CP) Module}
\label{ss_cpm}

The CP module generates control velocity commands to move the robot. Instead of just generating possible velocities to reach the next subgoal location, our module attempts to reach the $N$ next subgoal locations using the current image and the $N$ consecutive subgoal images. This allows our method to \textbf{avoid large obstacles} that could be placed on a subgoal location by transitioning to the next one.

The CP module is composed of $N$ units, each one considering a different subgoal image $\{ I_{s+i}\}_{i=1 \cdots N}$. These units have the same structure, and share the same weights. The units are recurrent: this allows them to generate multiple steps of virtual velocities (a trajectory) conditioned on the previous steps. The $i$-th unit in the CP module receives the $i$-th subgoal image $I_{s + i}$ and the current image $I_t$ and generates trajectories (in the form of temporal sequences of velocities) towards it. This design is equivalent to a flexible predictive planning horizon allowing to avoid obstacles in the visual path by reaching the next subgoals.

Instead of a single sequence of velocities towards a subgoal, each unit generates $M$ alternative sequences. These alternatives are sampled from a \textbf{probabilistic predictive model} that learns to map images to \textbf{possible paths} as demonstrated by human teleoperators in the datasets we use for training (see Sec.~\ref{s_training}). In this way, our model proposes different strategies to reach each subgoal image, as learned from human demonstrations. The entire CP module with $N$ units generates in total $N \times M$ sequences of velocities. 

The $j$-th generated velocity sequence for the $i$-th subgoal image $I_{s+i}$ is defined as $\nu_{i, j}=\{{u}_{i, j, k} \}_{k=0 \cdots 2L-1}$, and the sequence is $2L$ long. $u_{i, j, k}$ are the velocities commands, pairs of  linear and angular velocity, $(v_{i, j, k}, \omega_{i, j, k})$ for the wheeled robot. The sequence of $2L$ velocities are planned to be executed at a constant time interval $\Delta_t$.

The velocity commands $u_{i, j, k}$ are generated by iterative calculation of a Long short-term memory~(LSTM).
To generate $u_{i, j, k}$ we feeding into the LSTM unit the visual features extracted from the subgoal image and the current camera image, $z_i^{cp}= f_{cp}^{conv}(I_t, I_{s+i})$, concatenated with the previous output velocity command $u_{i, j, k-1}$. However, as explained before, in order to generate multiple possible trajectories, the output of the LSTM is not a single velocity command, but a probability distribution of possible commands. We assume the distribution to be Gaussian: the LSTM outputs mean and covariance of the distribution of possible next commands (linear and angular velocity) given previous command, current and subgoal images as $f_{cp}^{LSTM}(z_i^{cp}, u_{i, j, k-1}) = \mathcal{N}(\mu_{i, j, k}, \Sigma_{i, j, k})$.
%
%
%
%

Our method samples the distribution to get the linear and angular velocities as $u_{i, j, k} \sim \mathcal{N}(\mu_{i, j, k}, \Sigma_{i, j, k})$ for the $k$-th step.
%
%
We iterate these calculation for $2L$ times to obtain a trajectory of velocities, $\nu_{i, j}$, towards the $i$-th subgoal image.

The question now is how to train $f_{cp}^{conv}()$ and $f_{cp}^{LSTM}()$ to output feasible trajectories of velocities to reach each subgoal image. We propose to learn from human teleoperation commands the best strategies to navigate to subgoal based on images. However, given the high-dimensionality of the image inputs, the training process would benefit of additional supervision. We propose to learn to output velocity commands by optimizing \textbf{a combination of image loss, trajectory likelihood loss and smoothness loss}. While the trajectory likelihood loss is the common loss to learn to imitate demonstrations, the two other losses provide supervision on the image and velocity spaces.
Thus, during training, $f_{cp}^{conv}()$ and $f_{cp}^{LSTM}()$ are optimized by minimizing the following cost function, $J = J^{\textrm{\it bimg}} + \kappa_1 J^{\textrm{\it tl}} + \kappa_2 J^{\textrm{\it smo}}$, where $J^{\textrm{\it bimg}}$ is a bidirectional image loss, $J^{\textrm{\it tl}}$ is a trajectory likelihood loss, $J^{\textrm{\it smo}}$ is a smoothness loss, and $\kappa_1, \kappa_2$ are weighting factors to balance the objectives. We explain the loss design in the following paragraphs.
%
%
\subsubsection{Bidirectional Image Loss}
The basic strategy is to reach the subgoal location where it can obtain an image similar to the subgoal image. However, to avoid collisions, our method needs to be able to deviate from the original path. A longer predictive horizon may help to plan larger deviations for collision avoidance.
However, {\colornewparts the predictive performance deteriorates with a longer prediction horizon}, consequently hurting the search for optimal velocity and thus navigation performance, as shown in the experimental section.

We propose to apply the bidirectional image loss:
\begin{eqnarray}
\label{eq:jpixel2}
J^{\textrm{\it bimg}} &=& \sum_{i=0}^{N-1} \sum_{j=0}^{M-1} \left( \frac{1} {2L\cdot N_{\textrm{\it pix}}} \sum_{k=1}^L  w_k|\hat{I'}_{i, j, t+L} - \hat{I}_{i, j, t+k}| \right. \nonumber \\
&&+ \left. \sum_{k=L}^{2L-1} w_k|\hat{I}_{i, j, t+L} - \hat{I'}_{i, j, t+k}| \right),
\end{eqnarray}
where $N_{\textrm{\it pix}}$ denotes the total pixel count of the image and $\{\hat{I}_{i, j, t+k} \}_{k=1\cdots L}$ denotes predicted images conditioned on $\{u_{i, j, k}\}_{k=0 \cdots L-1}$ from the current camera image $I_t$. 
\begin{eqnarray}
\label{eq:vunet2_f}
\hspace{2mm}[\hat{I}_{i, j, t+1} \cdots \hat{I}_{i, j, t+L}] = f_{\mathit{VUNet}}(I_t, u_{i, j, 0}, \cdots, u_{i, j, L-1}).
\end{eqnarray}
Here, $f_{\mathit{VUNet}}()$ is the pretrained VUNet-360 neural network \cite{hirose2019vunet, hirose2019dvmpc}.
$\{ \hat{I'}_{i, j, t+k} \}_{k=L\cdots 2L-1}$ in (Eqn.~\ref{eq:jpixel2}) denote the images predicted using the subgoal image $I_{s+i}$, as follows:
\begin{eqnarray}
\label{eq:vunet2_b}
& & \hspace{-8mm}[\hat{I'}_{i, j, t+2L-1} \cdots \hat{I'}_{i, j, t+L}] \nonumber\\
& & \hspace{8mm} = f_{\mathit{VUNet}}(I_{s+i}, -u_{i, j, 2L-1}, \cdots, -u_{i, j, L}),
\end{eqnarray}
where the subgoal image $I_{s+i}$ is assumed to be the image $I_{t+2L}$ in (Eqn.~\ref{eq:vunet2_b}).
Note that the chronological order and signs of the velocities are inverted because the chronological order of the predicted images is also inverted in (Eqn.~\ref{eq:vunet2_b}). 
Using the subgoal image $I_{s+i}$ as $I_{t+2L}$, we can double the predictive horizon(= $2L$) without deterioration in image prediction
because the subgoal image $I_{s+i}$ can accept geometric and appearance information, which is not appeared in $I_t$.

\subsubsection{Trajectory Likelihood Loss}
The teleoperator's velocity commands recorded in the dataset contain rich information about navigating the robot while avoiding collisions, e.g. slowing down when going through narrow passages, slowing down when detouring around obstacles. Such rich, view-specific velocity commands can be leveraged in training navigation policies. 

To train the rich information in the teleoperator's velocity commands, our method applies the following trajectory likelihood loss, which is the negative log likelihood loss function defined over samples of teleoperator trajectory and our trajectory generation model. 
\begin{eqnarray}
\label{eq:cpl2}
&&J^{\textrm{\it tl}} = \sum_{i=0}^{N-1} \sum_{j=0}^{M-1} \left( \frac{1}{2L}\sum_{k=0}^{2L-1}\mbox{log} \sqrt{(2\pi)^3 |\bm{P}_{i, j, k}|} \right. \nonumber \\
&& \left. + \frac{1}{2}(\bm{x}^{ref}_{i, j, k} - \bar{\bm{x}}_{i, j, k})^T (\bm{P}_{i, j, k})^{-1} (\bm{x}^{ref}_{i, j, k} - \bar{\bm{x}}_{i, j, k})  \right),
\end{eqnarray}
where $\{ \bm{x}^{ref}_{i, j, k}\}_{k=1 \dots 2L}$ is the robot trajectory controlled by the teleoperator velocity $\{ u^{ref}_{i, j, k}\}_{k=0 \cdots 2L-1}$ and $\{ \bar{\bm{x}}_{i, j, k}\}_{k=1 \dots 2L}$ is the mean robot trajectory from generated velocity $\{ \mu_{i, j, k}\}_{k=0 \cdots 2L-1}$. Here, $\bar{\bm{x}}_{i, j, k}$ is calculated using the following equation and the mean of the generated velocity $\mu_{i, j, k-1}$.
\begin{eqnarray}
\label{eq:cpl3a}
\bar{\bm{x}}_{i, j, k} &=& \bm{x}_{i, j, k-1} + \left[
    \begin{array}{cc}
        \Delta_t \mbox{cos} \theta_{i, j, k-1} & 0.0 \\
        \Delta_t \mbox{sin} \theta_{i, j, k-1} & 0.0 \\
        0.0 & \Delta_t \\
    \end{array}
    \right] \mu_{i, j, k-1}\nonumber \\
    &=& f_{odom}(\bm{x}_{i, j, k-1}, \mu_{i, j, k-1}).
\end{eqnarray}
where $\{\bm{x}_{i, j, k}\}_{k=1 \dots 2L}$ is the virtual robot trajectory using the sampled velocity according to the distribution $\mathcal{N}(\mu_{i, j, k}, \Sigma_{i, j, k})$. 
The wheeled robot travels on a 2D plane. Hence, $\bar{\bm{x}}_{i, j, k}$ is defined by the robot position $\bar{x}_{i, j, k}, \bar{y}_{i, j, k}$ and yaw angle $\bar{\theta}_{i, j, k}$.
In addition, $\bm{x}_{i, j, k}$ and $\bm{x}^{ref}_{i, j, k}$ can be calculated as $\bm{x}_{i, j, k}=f_{odom}(\bm{x}_{i, j, k-1}, u_{i, j, k-1})$ and $\bm{x}^{ref}_{i, j, k}=f_{odom}(\bm{x}^{ref}_{i, j, k-1}, u^{ref}_{i, j, k-1})$, respectively.

The covariance matrix $\bm{P}_{i, j, k}$ of $\bm{\bar{x}}_{i, j, k}$ can be defined as $\bm{P}_{i, j, k} = \bm{F}_{i, j, k} \bm{\Sigma}_{i, j, k} \bm{F}_{i, j, k}^T + \bm{\Gamma}$, where $\bm{F}_{i, j, k}$ is the Jacobian of (Eqn.~\ref{eq:cpl3a}) for $\mu_{i, j, k-1}$, and $\bm{\Gamma}$ is the constant diagonal matrix.
%
By minimizing $J^{\textrm{\it tl}}$, the CP can generate velocity commands that roughly follow the distribution of the teleoperator command.

\subsubsection{Smoothness Loss}

Continuous velocity is required for the actual control of the robot so that the motion is not jerky. We use $J^{\textrm{\it smo}}$ as the objective to suppress discontinuous velocity commands.
$J^{\textrm{\it smo}}$ is designed as the sum of MSE between the linear and angular velocities of consecutive steps, $J^{smo} = \sum_{i=0}^{N-1} \sum_{j=0}^{M-1}\sum_{k=0}^{2L-2} (u_{i, j, k+1} - u_{i, j, k})^2$.

\subsection{Trajectory Selection~(TS) Module}
\label{ss_sm}

This module selects the best policy from the $N$ sets of sequences of velocities generated by the CP module to control the robot. In the TS module, $N$ units are used to evaluate the $N \times M$ velocities generated by the CP module. 

The TS module measures the score $S_{i, j}$ for all generated velocities and selects the minimum value $\nu_{i^\star, j^\star}$.
Here, $i^\star, j^\star$ is the optimal $i, j$ to minimize $S_{i, j} = S_{i, j}^{reac} + S_{i, j}^{trav}$.
$S_{i, j}^{reac}$ is a reachability score for reaching the subgoal, and $S_{i, j}^{trav}$ is a traversability score, which penalizes travelling in risky areas. In our method, a smaller score indicates a better path.

\subsubsection{Reachability Score}

To ensure the reachability to the subgoal location, the generated velocity vectors need to satisfy the condition $\hat{I}_{i, j, t+L} = \hat{I'}_{i, j, t+L}$, because we use $I_{s+i}$ as $I_{t+2L}$ in bidirectional prediction.
$S_{i, j}^{reac}$ is calculated as $S_{i, j}^{reac} = (f_{ev}(\hat{I}_{i, j, t+L})- f_{ev}(\hat{I'}_{i, j, t+L}))^2$ to measure the similarity of  $\hat{I}_{i, j, t+L}$ and $\hat{I'}_{i, j, t+L}$. $f_{ev}()$ is a neural network to extract the feature for calculation of $S_{i, j}^{reac}$.

%
%
%
\subsubsection{Traversability Score}
The velocity sequences corresponding to the untraversable path should be excluded as candidates to ensure the collision free motion.
In order to penalize the untraversable path, {\colornewparts we set $S_{i, j}^{trav} = P^{trav}$ if $\hat{p}^{trav}_{i, j , L} < 0.5$.
Otherwise, $S_{i, j}^{trav}$ is set as $0.0$. }
Here $\hat{p}^{trav}_{i, j, L}$ is the estimated traversable probability by feeding the representative predicted image $\hat{I}^f_{i, j, L}$ (namely, the front side of the predicted image at the $L$-th step from the current image view) into the neural network GONet \cite{hirose2018gonet}.
$P^{trav}$ must be substantially larger than the reachability score $S_{i, j}^{reac}$ to prioritize avoiding collision.

Hence, $S_{i, j}^{trav}$ can be used to effectively remove the untraversable trajectories.
We only choose one predicted image for the traversability estimation to reduce the computational load for online implementation.
In addition to the above selection process, we override the selected velocity with a pivot turning motion~(90$^\circ$ rotation) if the estimated traversable probability at the current robot pose is less than 0.5 to continue moving.
The sign of the pivot turning ($+$90$^\circ$ or $-$90$^\circ$) is decided by a sign of $\theta_{i^\star, j^\star, 2L}$ to try to face the next sub-goal location.
%
%
%
%
%

\subsection{Subgoal Selection~(SS) Module}
\label{ss_vlm}

The SS module selects the closest subgoal image $I_{s}$ from all subgoal images of the VT to feed the subsequent $N$ subgoal images $\{ I_{s+i}\}_{i=1 \cdots N}$ into the CP and TS modules.
One practical approach to decide $I_{s}$ is to measure the similarity between $I_t$ and all the subgoal images such as $S_{i, j}^{reac}$~\cite{pathak2018zero}.
However, we do not want to increase the computational load as the system needs to run in real-time. Comparing with each subgoal will increase the computational cost as the number of subgoals increases.

Instead of measuring the image similarity directly, we use the virtual velocities generated by the CP module to decide $s$ since it is more robust in practice.
Under the assumption that $\hat{I}_{i, j, t+L} = \hat{I'}_{i, j, t+L}$, the robot pose {\colornewparts $[x_{i, j, 2L}, y_{i, j, 2L}, \theta_{i, j, 2L}]^T$} at the 2$L$-th step estimated by $\nu_{i, j}$ approximately equals to the pose of the subgoal image on the robot coordinate.
{\colornewparts We provide the threshold conditions for the distance and yaw angle as $d_{th} > (x_{i, j, 2L}^2 + y_{i, j, 2L}^2)^{\frac{1}{2}}$ and $\theta_{th} > \theta_{i, j, 2L}$ for all $N \times M$ velocities, respectively.} 
{\colornewparts We select $i_{ss}$, which is the largest one in $i$ that satisfy the above inequalities and decide next $s$ as $s + i_{ss}$ to select the subgoal images close to the final goal. This process can allow the agent to transition to further subgoals even when the closest subgoals are occluded by large obstacles that are not present in the VT.}
\section{Training}
\label{s_training}

There are trainable elements (neural networks) in the CP module and in the TS module. To train them, we use the GO Stanford dataset 4 (GS4)~\cite{gostanford}.
GS4 was collected by teleoperating a TurtleBot2 mobile robot equipped with a RGB 360$^\circ$ camera.
The dataset contains synchronized video sequences from the 360$^\circ$ camera and teleoperator's commanded velocities when navigating in twelve buildings at the Stanford University campus for 10.3 hours.
We chose eight buildings for training, two buildings for validation, and two other buildings for testing. 
In addition to real world data, GS4 contains data from the Gibson Simulation Environment~\cite{xia2018gibson}, a photo-realistic simulator used to train navigation agents using models from Stanford 2D-3D-S~\cite{armeni2017joint} and Matterport3D~\cite{chang2017matterport3d}.
The simulator data in GS4 was collected by teleoperating a simulated robot with a virtual RGB 360$^\circ$ camera in 36 different reconstructed buildings for a total of 3.6 hours.
We separate 36 buildings to 26, 5, and 5 buildings for training, validation, and testing, respectively. 
When training the neural network models, we randomly mix the data from the real environment and the simulator to achieve better generalization.

\subsection{Training the CP Module}
\label{ss_tcpm}

To train $f_{cp}^{conv}()$ and $f_{cp}^{LSTM}()$ in the CP module, we randomly choose two images from the same trajectory of the dataset, $I_a$ and $I_b$, separated by $K$ steps ($K<2L$), so that $I_b = I_{a+K}$.
We assume the first image to be the current image, $I_t=I_a$ and the second image to be each of the subsequent subgoals, ${I}_{s+i} = I_b$ for $i\in\{0\ldots N-1\}$. We resize the image to 128 $\times$ 128 pixels and feed them into $f_{cp}^{conv}()$ to generate a feature, $z_i^{cp}$. 
We randomly sample $u_{i, j, k}$ with $i=j=1$ according to the Gaussian distribution $\mathcal{N}(\mu_{i, j, k}, \Sigma_{i, j, k})$ during the training.

$\{ u_{i, j, k}^{ref} \}_{k=0 \cdots 2L-1}$ is provided by the teleoperator's velocity to calculate the trajectory likelihood loss $J^{\textrm{\it prob}}$. 
We set 0.0 for $\{ u_{i, j, k}^{ref} \}_{k=0 \cdots 2L-1}$ if $k \geq K$.
Moreover, the initial pose, $\bar{\bm{x}}_{i, j, -1}$, $\bm{x}_{i, j, -1}$, and $\bm{x}_{i, j, -1}^{ref}$ are defined as zeros to calculate the trajectory on the robot coordinate at $k=0$.

\subsection{Training the TS Module}

$f_{ev}()$ is the neural network with seven convolutional layers using batch normalization and the leaky ReLU function to extract the feature, and $S_{i, j}^{reac}$ should take a bigger value when the Euclidean distance between the corresponding predicted images is bigger.
We train $f_{ev}()$ using triplet loss \cite{wang2014learning}.
First, we calculate the trained CP module in the same manner as that used for training the CP module.
We generate $2L$ predicted images conditioned on the obtained virtual velocities using (Eqn.~\ref{eq:vunet2_f}) and (Eqn.~\ref{eq:vunet2_b}).
Then, we randomly choose one anchor image and two images from the $2L$ predicted images.
One of the images, which is closer to the anchor image on the Euclidean coordinate of robot position, is set as the positive image, and another is set as the negative image to calculate the triplet loss.
Note that the Euclidean distance of robot position between two predicted images is calculated by using the virtual velocities generated by the CP module.
By iteratively updating $f_{ev}()$ to minimize the triplet loss, we train $f_{ev}()$ to extract the feature for $S_{i, j}^{reac}$.
\subsection{Parameters}
The horizon $L$ is set to 8, which is same as one of our baselines \cite{hirose2019dvmpc}, for a fair comparison.
Hence, the control and predictive horizons of our method are 5.328 s ($=2L\times \Delta_t = 16 \times 0.333$).
The maximum predictive range can be 2.666 m and  $\pm$ 5.328 rad, because the maximum linear and angular velocities are designed to be 0.5 m/s and 1.0 rad/s, respectively.
In $J^{\textrm{\it bimg}}$, the weighting factor $\{w_k \}_{k=1 \cdots 2L-1}$ is set as 1.0 except when $k=L$.
$w_L = 5.0$ is used to give the soft constraint for $\hat{I}_{t+L} = \hat{I'}_{t+L}$.
The weighting factors to balance each objective in $J$ are designed as $\kappa_1 = 0.001$ and $\kappa_2 = 1.0$ after some trial and error.

For the TS module, $P^{trav}$ for $S_{i, j}^{trav}$ is designed as 100.0 ($\gg S_{i, j}^{reac}$) to penalize the untraversable velocity sequence.
Threshold values $d_{th}$ and $\theta_{th}$ in the SS module are set as 0.8 m and 0.4 rad, respectively.
Larger $N$ and $M$ can improve the robustness of reaching the position of the subgoal image and avoiding collisions, because we can check the reachability and traversability for more cases ($= N \times M$ cases).
However, the calculation cost will be higher for larger $N$ and $M$.
Hence, both $N$ and $M$ take the design value of 3 at the inference time although $N = M = 1$ is used for training.
We select three velocity vectors from $\mathcal{N}(\mu_{i, j, k}, \Sigma_{i, j, k})$ by repeatedly selecting the mean point and the $\pm$ 2 sigma point for 2$L$ steps. 
%
\section{Experiments}
\label{s_exp}

{\colornewparts We conduct experiments to evaluate our method in simulated and real environments.}
First, we validate our method by comparing its navigation results in simulation against those of state-of-the-art baseline methods that learn to follow subgoal images in a visual navigation.
Next, we test our method and the baselines on a real physical mobile robot, including one different to the one used to collect the training dataset. With this test we evaluate the generalizability and robustness of our method among different embodiments. 
We also include a qualitative and visual analysis of the trajectories followed to provide insights of the behavior of our method.
In all our experiments, we evaluate the capacity of the navigation methods to move along a VT acquired before via teleoperation (in simulation or real world), and whether they can deviate and come back to the trajectory when there is an obstacle in the path that was not present during the collection of the VT.

\begin{table}[t]
  \vspace*{1.5mm}
  \caption{{\small{\bf Comparison with baseline methods}. Navigation results with our method and the baseline methods in 10 environments and 1000 trials each (Table shows mean of goal arrival rate / subgoal coverate rate / success weighted by path length (SPL)\cite{anderson2018evaluation})}}
  \vspace*{-3mm}
  \begin{center}
  \resizebox{1.0\columnwidth}{!}{
  \label{tab:ev}
  \begin{tabular}{lc|c|c} \hline
    method & step & \hspace{15mm}results\hspace{15mm} & Hz \\ \hline
    open loop & $-$ & 0.016~~/~~0.371~~/~~0.016 & $-$ \\ \hline
    Imitation learning (IL) & 8 & 0.346~~/~~0.695~~/~~0.342 & 13.703 \\ 
    & 16 & 0.208~~/~~0.632~~/~~0.196 & 10.068 \\ \hline
    Zero-shot visual imitation (ZVI) \cite{pathak2018zero} & 8 & 0.370~~/~~0.695~~/~~0.364 & 13.726 \\ 
    & 16 & 0.127~~/~~0.562~~/~~0.102 & 10.059 \\ \hline
    Stochastic optimization (SO) \cite{finn2017deep} & 8 &  0.611~~/~~0.793~~/~~0.610 & 0.585 \\ 
    {\it NOT calculated in real time} & 16 &  0.467~~/~~0.734~~/~~0.466 & 0.341 \\ \hline
    Deep visual MPC (DVMPC) \cite{hirose2019dvmpc} & 8 & 0.700~~/~~0.850~~/~~0.700 & 13.498 \\ 
    & 16 & 0.469~~/~~0.765~~/~~0.468 & 9.967 \\ \hline
    {\bf Our method~({\it full})} & 16 & {\bf 0.807}~~/~~{\bf 0.905}~~/~~{\bf 0.803} & 5.342 \\
    {\it \hspace{2mm} w/o bidirectional prediction} & 8 & 0.777~~/~~0.892~~/~~0.784 & 7.321 \\
    {\it \hspace{2mm} w/o probabilistic control} & 16 & 0.731~~/~~0.870~~/~~0.729 & 9.827 \\
    {\it \hspace{2mm} w/o our subgoal selection} & 16 & 0.787~~/~~0.889~~/~~0.784 & 5.258 \\ \hline
  \end{tabular}
  }
  \end{center}
  \vspace*{-5mm}
\end{table}
\begin{figure}[t]
  \begin{center}
  \hspace*{-5mm}
      \includegraphics[width=0.8\hsize]{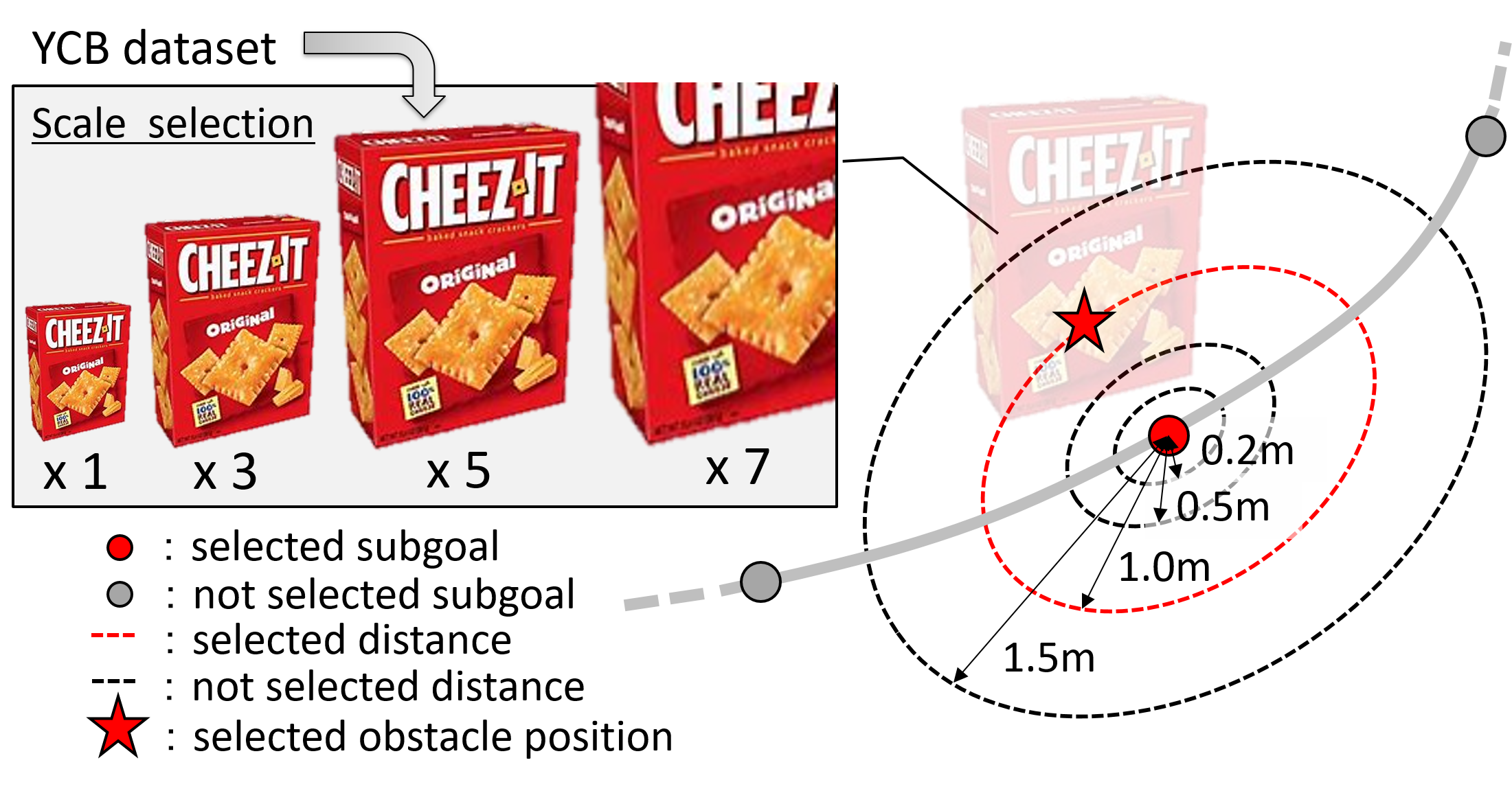}
  \end{center}
      \vspace*{-3mm}
	\caption{\small {\bf Overview of obstacle placement in simulation.} Scale of object, corresponding subgoal, and distance from the subgoal position are randomly selected to place the object in each simulation. The obstacle position~(red star) is randomly decided on the selected red circle.}
  \label{f:obstacle}
  \vspace*{-7mm}
\end{figure}
\subsection{Comparison to Baselines in Simulation}
\label{ss_ctb}
First, we compare the navigation performance of our method to the performance of state-of-the-art baselines. 

For each method, we evaluate the performance in the same 100 trajectories in 10 different environments and measure the goal arrival rate, subgoal coverage rate, and success weighted by path length (SPL)~\cite{anderson2018evaluation}.
We define the goal region as a circle of \SI{0.5}{\meter} around the goal location.

We evaluate the navigation performance of our method and baselines to avoid an obstacle placed in the visual path. We randomly choose meshes from the YCB dataset~\cite{calli2017yale} and scale it to 1, 3, 5, or 7 times the original size on all dimensions. If the agent collides with the obstacle or with the environment, we stop the trial and consider it finished (failed).  
To place the obstacle we randomly select a subgoal location in the VT, excluding the final subgoal or subgoals in very narrow areas, e.g., corridors or doors. Obstacles in very narrow areas would make impossible to navigate and reach the goal.
Then, we pick a distance randomly among \SI{0.2}{\meter}, \SI{0.5}{\meter}, \SI{1.0}{\meter}, or \SI{1.5}{\meter} and randomly place the obstacle on the circle whose radius is selected distance as shown in Fig.~\ref{f:obstacle}. 
The robot does not start at the exact initial location of the VT, but a distance to it chosen randomly between \SI{0}{\meter}, \SI{0.1}{\meter}, \SI{0.2}{\meter}, and \SI{0.3}{\meter}. We simulate realistic random noise in the controller execution by adding slippage. Per trajectory, we select randomly a slippage ratio between 0.0, 0.1, 0.2 and 0.3, that indicate the ratio of motion that is lost due to slippage. For example, a ratio of 0.1 implies that if the agent commands $u=(v,\omega)$, the environment will only execute $(1-0.1)u$. This emulates different types of floors.

We compare our method to the following five baselines:
\begin{enumerate}[leftmargin=*]
\item \textbf{Open loop control:}
The robot moves by directly replaying the teleoperator's velocities in the simulator. 
\item \textbf{Imitation Learning (IL):}
We train a convolutional neural network with the same structure and size as the feature extraction network ($f_{cp}^{conv}$) of our method to map images to teleoperator's velocities by minimizing MSE (behavioral cloning). 
\item \textbf{Zero-shot Visual Imitation (ZVI) \cite{pathak2018zero}:}
Similar to our method, ZVI trains the control policy in an offline fashion using the predictive model, and implements the trained model to obtain the robot command.
\item \textbf{Stochastic Optimization (SO) \cite{finn2017deep}:}
We develop a cross-entropy method (CEM), which is a stochastic optimization algorithm similar to the one used by \citet{finn2017deep}.
First, we sample 50 linear and angular velocity vectors and compute the objectives for DVMPC~\cite{hirose2019dvmpc}.
Then, we select the 5 velocities with lowest cost and resample a new set of 50 velocities from a multivariate Gaussian distribution.
We iterate 5 times the above process and query the velocities with the lowest cost.
\item \textbf{Deep Visual Model Predictive Control (DVMPC) \cite{hirose2019dvmpc}:}
\citet{hirose2019dvmpc} trained offline a deterministic control policy to minimize a MPC objective with not bidirectional predictive model, including a penalization for collision avoidance.
This process is less computationally expensive than the online optimization process~\cite{finn2017deep}.
\end{enumerate}
To remove the advantage in our method provided by using a panoramic 360$^\circ$ camera, we retrain all the baselines using also 360$^\circ$ images from the GS4 dataset~\cite{gostanford}.

{\colornewparts In simulation, we evaluate our method~(full) and our method excluding bidirectional prediction, probabilistic control approach, or our subgoal selection to ablate their benefits. 
Our method without the bidirectional prediction applies the unidirectional image loss of \cite{hirose2019dvmpc}, instead of $J^{bimg}$.
We set the predictive horizon as 8$(=L)$, because longer horizon~($2L$) degrades the performance caused by not using bidirectional prediction as shown in Table \ref{tab:ev}.
For our method without the probabilistic control approach, we apply the MSE loss to deterministically mimic the teleoperator's reference command, instead of $J^{\textrm{\it tl}}$. And, we measure the image similarity between the current image and the subgoal images for the subgoal selection, according to \cite{hirose2019dvmpc}.}

Table \ref{tab:ev} shows the evaluation results for the baseline methods and the proposed method in simulation, including the action output frequency.
We observe that a longer predictive horizon (16) degrades the performance of the baselines.
This drop results from the increased difficulty of predicting far future commands due to occlusions and unforeseeable effects. 
Differently, our method {\colornewparts (full)} shows the best performance of all approaches, indicating that the bidirectional prediction approach degrades less than unidirectional predictions.

\begin{figure}[t]
  \begin{center}
  \vspace*{1.5mm}
    \includegraphics[width=0.99\hsize]{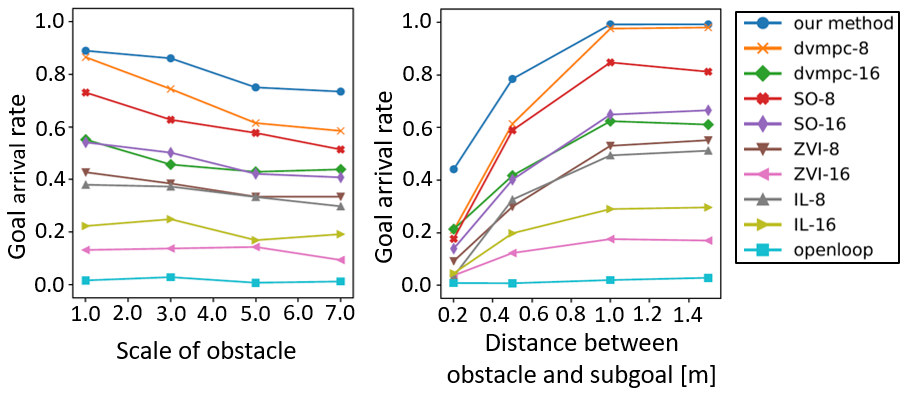}
  \end{center}
    \vspace*{-3mm}
    \caption{\small {\bf Navigation performance wrt. size and location of obstacles:} Goal arrival rate as a function of a) the scale of the obstacle and b) distance between the center of the obstacle and the VT. Our method is more robust and can deviate to avoid larger obstacles and obstacles close to the subgoal.}
  \label{f:ev_result}
\end{figure}

Fig.~\ref{f:ev_result} depicts the performance (goal arrival rate) of the methods with respect to the size of the obstacles (left) and the distance to the VT (right).
As expected, all the baseline methods decrease their success rate for larger obstacle scale and lower distance to the trajectory.
However, compare to all the baselines, the performance of our method degrades less. In the most difficult cases, with maximum obstacle scale and shortest distance to the VT our method achieves around 15 to 20\% better results than the second best baseline method.

\subsection{Real Robot Experiments}

We evaluate our method in the real world with a physical robot.
We implement our method on a laptop with Nvidia GeForce GTX 1070 and mount it on TurtleBot2 with a 360$^\circ$ camera (Ricoh THETA S).
We compare our method to the baselines defined in the previous section. 
In our evaluation we include dynamic obstacles, pedestrians.
In each environment, we conduct 10 trials by varying the obstacle position, initial robot position, lighting condition, and pedestrian positions.
Table \ref{tab:renv} shows the mean goal arrival rate and subgoal coverage rate in three different environments.
Some example trajectories of our method are included in our supplemental video.
We observe the same behavior as in simulation: our method results in a better goal arrival rate than the strongest baseline approach, DVMPC with the 8-step horizon.

\begin{table}[t]
  \centering
  \caption{{\small{\bf Navigation performance in a real environment} (goal arrival rate/subgoal coverage rate)} The distance besides the case number is a length of VT.}
  \label{tab:renv}
  \resizebox{0.85\columnwidth}{!}{
  \begin{tabular}{l|c|c|c}\cline{1-4}
       & Case 1: 8.1 m & Case 2: 13.7 m & Case 3: 9.5 m \\ \cline{1-4}
       DVMPC (8 steps) & 0.500 \,/ \,0.775 & 0.500 \,/ \,0.688 & 0.600 \,/ \, 0.827 \\ \cline{1-4}
       Our method & 0.800 \,/ \,0.892 & 0.900 \,/ \,0.944 & 0.800 \,/ \, 0.918 \\ \cline{1-4}
  \end{tabular}
  }
  \vspace{-1.5em}
\end{table}

Finally, we experiment on a different robot platform. 
The data collection and real experiments shown above are performed on a Turtlebot 2. 
To validate the robustness to different robot embodiments, we mount the same laptop PC on a different built in-house mobile robot (shown in supplementary video). 
Compared to Turtlebot 2, this robot has a different mechanical structure, inertia, weight, and sensor position.
We conduct 10 trials in 2 environments using this robot.
Our method arrives at the goal region with a success rate of 80\%.

\subsection{Qualitative Analysis}

\begin{figure}[t]
  \vspace{2.5mm}
  \begin{center}
    \includegraphics[width=0.99\hsize]{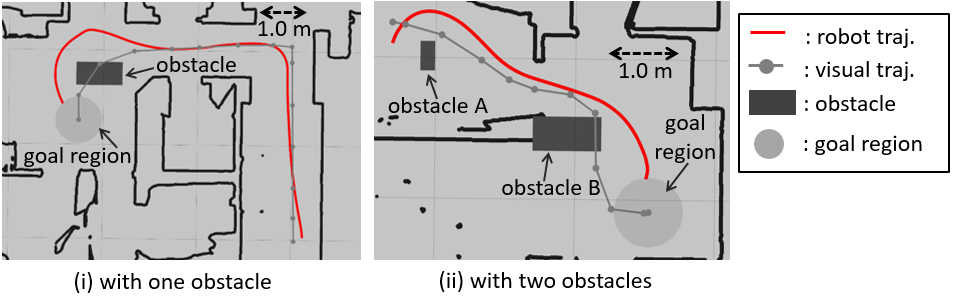} \\
  \end{center}
    \vspace*{-3mm}
	\caption{\small {\bf Visualization of our method} in top-down view of the robot. The robot trajectories (red line) deviates from the original tele-operated trajectory (gray line) to avoid the obstacles (dark gray rectangles) and returns successfully to the path to reach the goal.}
	\vspace*{-5mm}
  \label{f:ex_dist}
\end{figure}

%
%
%
Fig.~\ref{f:ex_dist} depicts trajectories of the robot as executed by our method.
In the presence of one large obstacle occludes the view of the next subgoal (left), our method follows the VT in the first part and deviates from it in the final part from the VT to avoid the collision. 
Even when two obstacles are placed on the path (right), our method allows the robot to avoid them and reach the goal destination.
{\colornewparts Additional results of qualitative analysis are shown in the supplemental video. }
\section{Conclusion and Future work} 
\label{sec:conclusion}

We presented a novel control system to allow a mobile robot to navigate using only one 360$^\circ$ fisheye RGB camera.
Our work makes three main contributions: 1) A novel bidirectional image prediction method that extends the image predicting horizon, 2) A probabilistic representation of velocities that encodes and imitates non-deterministic human demonstrations, and 3) A robust system that combines these componets that allows the robot to navigate towards a goal, following a sequence of images even under significant changes between the VT and the current environmental conditions. 
Our experiments indicated that our method follows the VT robustly, avoiding unseen large obstacles in both simulated and real environments.
We also showed that our method applies to different mobile robot, even when they were not used to collect the human demonstrations of our training dataset.
However, our method still suffers from some limitations. Firstly, the robot collides occasionally because of GONet~\cite{hirose2018gonet}, the network that assesses traversability occasionally generate wrong predictions. {\colornewparts Secondly, the robot sometimes loses its own location when ``kidnapped" to a new position, causing SS module to fail, so integration with global mapping and planning methods will be required~\cite{savinov2018semiparametric,chen2019behavioral,meng2019scaling,chaplot2020neural} to solve this type of failures.} 
\balance
\bibliographystyle{IEEEtranN}
\vskip-\parskip
\begingroup
\footnotesize
\bibliography{reference_RAL}

\begin{thebibliography}{40}
\providecommand{\natexlab}[1]{#1}
\providecommand{\url}[1]{#1}
\csname url@samestyle\endcsname
\providecommand{\newblock}{\relax}
\providecommand{\bibinfo}[2]{#2}
\providecommand{\BIBentrySTDinterwordspacing}{\spaceskip=0pt\relax}
\providecommand{\BIBentryALTinterwordstretchfactor}{4}
\providecommand{\BIBentryALTinterwordspacing}{\spaceskip=\fontdimen2\font plus
\BIBentryALTinterwordstretchfactor\fontdimen3\font minus
  \fontdimen4\font\relax}
\providecommand{\BIBforeignlanguage}[2]{{%
\expandafter\ifx\csname l@#1\endcsname\relax
\typeout{** WARNING: IEEEtranN.bst: No hyphenation pattern has been}%
\typeout{** loaded for the language `#1'. Using the pattern for}%
\typeout{** the default language instead.}%
\else
\language=\csname l@#1\endcsname
\fi
#2}}
\providecommand{\BIBdecl}{\relax}
\BIBdecl

\bibitem[Chaumette and Hutchinson(2006)]{chaumette2006visual}
F.~Chaumette and S.~Hutchinson, ``Visual servo control. i. basic approaches,''
  \emph{IEEE RA Magazine}, vol.~13, no.~4, pp. 82--90, 2006.

\bibitem[Chaumette and Hutchinson(2007)]{chaumette2007visual}
------, ``Visual servo control. ii. advanced approaches [tutorial],''
  \emph{IEEE RA Magazine}, vol.~14, no.~1, pp. 109--118, 2007.

\bibitem[Kumar et~al.(2018)]{NIPS2018_7357}
A.~Kumar \emph{et~al.}, ``Visual memory for robust path following,'' in
  \emph{NeurIPS}, 2018, pp. 765--774.

\bibitem[Wang and Spelke(2002)]{wang2002human}
R.~F. Wang and E.~S. Spelke, ``Human spatial representation: Insights from
  animals,'' \emph{Trends in cognitive sciences}, vol.~6, no.~9, pp. 376--382,
  2002.

\bibitem[Chersi et~al.(2013)]{chersi2013mental}
F.~Chersi \emph{et~al.}, ``Mental imagery in the navigation domain: a
  computational model of sensory-motor simulation mechanisms,'' \emph{Adaptive
  Behavior}, vol.~21, no.~4, pp. 251--262, 2013.

\bibitem[Arnold et~al.(2016)]{arnold2016mental}
A.~E. Arnold \emph{et~al.}, ``Mental simulation of routes during navigation
  involves adaptive temporal compression,'' \emph{Cognition}, vol. 157, pp.
  14--23, 2016.

\bibitem[Drews et~al.(2017)]{pmlr-v78-drews17a}
P.~Drews \emph{et~al.}, ``Aggressive deep driving: Combining convolutional
  neural networks and model predictive control,'' in \emph{Proceedings of
  CoRL}, 2017, pp. 133--142.

\bibitem[Drews et~al.(2019)]{drews2019vision}
------, ``Vision-based high-speed driving with a deep dynamic observer,''
  \emph{IEEE RA-L}, vol.~4, no.~2, pp. 1564--1571, 2019.

\bibitem[Hirose et~al.(2019{\natexlab{a}})]{hirose2019dvmpc}
N.~Hirose \emph{et~al.}, ``Deep visual mpc-policy learning for navigation,''
  \emph{IEEE RA-L}, vol.~4, no.~4, pp. 3184--3191, 2019.

\bibitem[Finn and Levine(2017)]{finn2017deep}
C.~Finn and S.~Levine, ``Deep visual foresight for planning robot motion,'' in
  \emph{ICRA}.\hskip 1em plus 0.5em minus 0.4em\relax IEEE, 2017, pp.
  2786--2793.

\bibitem[Kase et~al.(2019)]{kase2019learning}
K.~Kase \emph{et~al.}, ``Learning multiple sensorimotor units to complete
  compound tasks using an rnn with multiple attractors,'' in \emph{IROS}.\hskip
  1em plus 0.5em minus 0.4em\relax IEEE, 2019.

\bibitem[Ebert et~al.(2018)]{ebert2018visual}
F.~Ebert \emph{et~al.}, ``Visual foresight: Model-based deep reinforcement
  learning for vision-based robotic control,'' \emph{arXiv preprint
  arXiv:1812.00568}, 2018.

\bibitem[LaValle(2006)]{lavalle2006planning}
S.~M. LaValle, \emph{Planning algorithms}.\hskip 1em plus 0.5em minus
  0.4em\relax Cambridge university press, 2006.

\bibitem[LaValle and Kuffner~Jr(2001)]{lavalle2001randomized}
S.~M. LaValle and J.~J. Kuffner~Jr, ``Randomized kinodynamic planning,''
  \emph{The international journal of robotics research}, vol.~20, no.~5, pp.
  378--400, 2001.

\bibitem[Hirose et~al.(2018)]{hirose2018gonet}
N.~Hirose \emph{et~al.}, ``Gonet: A semi-supervised deep learning approach for
  traversability estimation,'' in \emph{IROS}.\hskip 1em plus 0.5em minus
  0.4em\relax IEEE, 2018, pp. 3044--3051.

\bibitem[Hutchinson et~al.(1996)]{hutchinson1996tutorial}
S.~Hutchinson \emph{et~al.}, ``A tutorial on visual servo control,'' \emph{IEEE
  trans. on RA}, vol.~12, no.~5, pp. 651--670, 1996.

\bibitem[Mur-Artal and Tard{\'o}s(2017)]{mur2017orb}
R.~Mur-Artal and J.~D. Tard{\'o}s, ``Orb-slam2: An open-source slam system for
  monocular, stereo, and rgb-d cameras,'' \emph{IEEE Trans. on RO}, vol.~33,
  no.~5, pp. 1255--1262, 2017.

\bibitem[Kim and Eustice(2013)]{kim2013perception}
A.~Kim and R.~M. Eustice, ``Perception-driven navigation: Active visual slam
  for robotic area coverage,'' in \emph{ICRA}.\hskip 1em plus 0.5em minus
  0.4em\relax IEEE, 2013, pp. 3196--3203.

\bibitem[Karlsson et~al.(2005)]{karlsson2005vslam}
N.~Karlsson \emph{et~al.}, ``The vslam algorithm for robust localization and
  mapping,'' in \emph{ICRA}.\hskip 1em plus 0.5em minus 0.4em\relax IEEE, 2005,
  pp. 24--29.

\bibitem[Savinov et~al.(2018)]{savinov2018semiparametric}
N.~Savinov \emph{et~al.}, ``Semi-parametric topological memory for
  navigation,'' in \emph{ICLR}, 2018.

\bibitem[Chen et~al.(2019)]{chen2019behavioral}
K.~Chen \emph{et~al.}, ``A behavioral approach to visual navigation with graph
  localization networks,'' \emph{arXiv preprint arXiv:1903.00445}, 2019.

\bibitem[Meng et~al.(2019)]{meng2019scaling}
X.~Meng \emph{et~al.}, ``Scaling local control to large-scale topological
  navigation,'' \emph{arXiv preprint arXiv:1909.12329}, 2019.

\bibitem[Chaplot et~al.(2020)]{chaplot2020neural}
D.~S. Chaplot \emph{et~al.}, ``Neural topological slam for visual navigation,''
  in \emph{CVPR}, 2020.

\bibitem[Wijmans et~al.(2020)]{Wijmans2020DD-PPO:}
E.~Wijmans \emph{et~al.}, ``Dd-ppo: Learning near-perfect pointgoal navigators
  from 2.5 billion frames,'' in \emph{ICLR}, 2020.

\bibitem[Mishkin et~al.(2019)]{mishkin2019benchmarking}
D.~Mishkin \emph{et~al.}, ``Benchmarking classic and learned navigation in
  complex 3d environments,'' \emph{arXiv preprint arXiv:1901.10915}, 2019.

\bibitem[Zhu et~al.(2017)]{zhu2017target}
Y.~Zhu \emph{et~al.}, ``Target-driven visual navigation in indoor scenes using
  deep reinforcement learning,'' in \emph{ICRA}.\hskip 1em plus 0.5em minus
  0.4em\relax IEEE, 2017, pp. 3357--3364.

\bibitem[Kahn et~al.(2018)]{kahn2018self}
G.~Kahn \emph{et~al.}, ``Self-supervised deep reinforcement learning with
  generalized computation graphs for robot navigation,'' in \emph{ICRA}.\hskip
  1em plus 0.5em minus 0.4em\relax IEEE, 2018, pp. 1--8.

\bibitem[Codevilla et~al.(2018)]{codevilla2018end}
F.~Codevilla \emph{et~al.}, ``End-to-end driving via conditional imitation
  learning,'' in \emph{ICRA}.\hskip 1em plus 0.5em minus 0.4em\relax IEEE,
  2018, pp. 1--9.

\bibitem[Pokle et~al.(2019)]{pokle2019deep}
A.~Pokle \emph{et~al.}, ``Deep local trajectory replanning and control for
  robot navigation,'' in \emph{ICRA}.\hskip 1em plus 0.5em minus 0.4em\relax
  IEEE, 2019, pp. 5815--5822.

\bibitem[Pathak et~al.(2018)]{pathak2018zero}
D.~Pathak \emph{et~al.}, ``Zero-shot visual imitation,'' in \emph{Proceedings
  of CVPR Workshops}, 2018, pp. 2050--2053.

\bibitem[Amini et~al.(2019)]{amini2019variational}
A.~Amini \emph{et~al.}, ``Variational end-to-end navigation and localization,''
  in \emph{ICRA}.\hskip 1em plus 0.5em minus 0.4em\relax IEEE, 2019, pp.
  8958--8964.

\bibitem[Bansal et~al.(2019)]{bansal2019-lb-wayptnav}
S.~Bansal \emph{et~al.}, ``Combining optimal control and learning for visual
  navigation in novel environments,'' in \emph{Proceedings of CoRL}, 2019.

\bibitem[Hirose et~al.(2019{\natexlab{b}})]{hirose2019vunet}
N.~Hirose \emph{et~al.}, ``Vunet: Dynamic scene view synthesis for
  traversability estimation using an rgb camera,'' \emph{IEEE RA-L}, vol.~4,
  no.~2, pp. 2062--2069, 2019.

\bibitem[gos()]{gostanford}
``Go stanford dataset 4,''
  \url{http://svl.stanford.edu/projects/dvmpc/dataset/}, accessed: 2020-01-31.

\bibitem[Xia et~al.(2018)]{xia2018gibson}
F.~Xia \emph{et~al.}, ``Gibson env: Real-world perception for embodied
  agents,'' in \emph{Proceedings of CVPR}, 2018, pp. 9068--9079.

\bibitem[Armeni et~al.(2017)]{armeni2017joint}
I.~Armeni \emph{et~al.}, ``Joint 2d-3d-semantic data for indoor scene
  understanding,'' \emph{arXiv preprint arXiv:1702.01105}, 2017.

\bibitem[Chang et~al.(2017)]{chang2017matterport3d}
A.~Chang \emph{et~al.}, ``Matterport3d: Learning from rgb-d data in indoor
  environments,'' in \emph{3DV}.\hskip 1em plus 0.5em minus 0.4em\relax IEEE,
  2017, pp. 667--676.

\bibitem[Wang et~al.(2014)]{wang2014learning}
J.~Wang \emph{et~al.}, ``Learning fine-grained image similarity with deep
  ranking,'' in \emph{Proceedings of CVPR}, 2014, pp. 1386--1393.

\bibitem[Anderson et~al.(2018)]{anderson2018evaluation}
P.~Anderson \emph{et~al.}, ``On evaluation of embodied navigation agents,''
  \emph{arXiv preprint arXiv:1807.06757}, 2018.

\bibitem[Calli et~al.(2017)]{calli2017yale}
B.~Calli \emph{et~al.}, ``Yale-cmu-berkeley dataset for robotic manipulation
  research,'' \emph{The International Journal of Robotics Research}, vol.~36,
  no.~3, pp. 261--268, 2017.

\end{thebibliography}
\endgroup
%
%
%
\end{document}